\documentclass[runningheads]{llncs}
\usepackage{graphicx}

\usepackage{tikz}
\usepackage{comment}
\usepackage{amsmath,amssymb} 
\usepackage{color}

\usepackage{graphicx}
\usepackage{amsmath,amssymb} 
\usepackage{color}
\usepackage{epstopdf}
\usepackage{multirow}
\usepackage{authblk}

\usepackage{subfigure}

\begin{document}
\pagestyle{headings}
\mainmatter
\def\ECCVSubNumber{3331}  

\title{Video-based Remote Physiological Measurement via Cross-verified Feature Disentangling} 

\titlerunning{Remote Physiological Measurement via Cross-verified Disentangling}
%
\author{Xuesong Niu\inst{1,2} \and Zitong Yu\inst{3} \and Hu Han\inst{1,4} \and Xiaobai Li\inst{3} \and Shiguang Shan\inst{1,2,4} \and Guoying Zhao\inst{3}}
\authorrunning{X. Niu et al.}
%
\institute{Key Laboratory of Intelligent Information Processing of Chinese Academy of Sciences (CAS), Institute of Computing Technology, CAS, Beijing 100190, China.
\and
University of Chinese Academy of Sciences, Beijing, 100049, China
\and
Center for Machine Vision and Signal Analysis, University of Oulu, Finland
\and
Peng Cheng Laboratory, Shenzhen, 518055, China\\
\email{xuesong.niu@vipl.ict.ac.cn, }
\email{\{hanhu, sgshan\}@ict.ac.cn, }\\
\email{\{zitong.yu, xiaobai.li, guoying.zhao\}@oulu.fi}}
\maketitle

\begin{abstract}
Remote physiological measurements, e.g., remote photoplethysmography (rPPG) based heart rate (HR), heart rate variability (HRV) and respiration frequency (RF) measuring, are playing more and more important roles under the application scenarios where contact measurement is inconvenient or impossible. Since the amplitude of the physiological signals is very small, they can be easily affected by head movements, lighting conditions, and sensor diversities. To address these challenges, we propose a cross-verified feature disentangling strategy to disentangle the physiological features with non-physiological representations, and then use the distilled physiological features for robust multi-task physiological measurements. We first transform the input face videos into a multi-scale spatial-temporal map (MSTmap), which can suppress the irrelevant background and noise features while retaining most of the temporal characteristics of the periodic physiological signals. Then we take pairwise MSTmaps as inputs to an autoencoder architecture with two encoders (one for physiological signals and the other for non-physiological information) and use a cross-verified scheme to obtain physiological features disentangled with the non-physiological features. The disentangled features are finally used for the joint prediction of multiple physiological signals like average HR values and rPPG signals. Comprehensive experiments on different large-scale public datasets of multiple physiological measurement tasks as well as the cross-database testing demonstrate the robustness of our approach.
\end{abstract}

\section{Introduction}

Physiological signals, such as average heart rate (HR), respiration frequency (RF) and heart rate variability (HRV), are very important for cardiovascular activity analysis and have been widely used in healthcare applications. Traditional HR, RF and HRV measurements are based on the electrocardiography (ECG) and contact photoplethysmography (cPPG) signals, which require dedicated skin-contact devices for data collection. The usage of these contact sensors may cause discomfort and inconvenience for subjects in healthcare scenarios. Recently, a growing number of physiological measurement techniques have been proposed based on remote photoplethysmography (rPPG) signals, which could be captured from face by ordinary cameras and without any contact. These techniques make it possible to measure HR, RF and HRV remotely and have been developed rapidly~\cite{poh2010non,poh2011advancements,de2013robust,li2014remote,Tulyakov2016Self,wang2017algorithmic,niu2019rhythmnet,yu2019ICCV}.

The rPPG based physiological measurement is based on the fact that optical absorption of a local tissue varies periodically with the blood volume, which changes accordingly with the heartbeats. However, the amplitude of the light absorption variation w.r.t. blood volume pulse (BVP) is very small, i.e., not visible for human eyes, and thus it can be easily affected by factors such as subject's movements and illumination conditions. To solve this problem, many traditional methods are proposed to remove the non-physiological information using color space projection~\cite{de2013robust,wang2017algorithmic} or signal decomposition~\cite{poh2010non,poh2011advancements,Tulyakov2016Self}. However, these methods are based on assumptions such as linear combination assumption~\cite{poh2010non,poh2011advancements} or certain skin reflection models~\cite{de2013robust,wang2017algorithmic}, which may not hold in less-constrained scenarios with large head movement or dim lighting condition.

Besides the hand-crafted traditional methods, there are also some approaches which adopt the strong modeling ability of deep neural networks for remote physiological signal estimation~\cite{spetlik2018BMVC,chen2018deepphys,niu2019rhythmnet,yu2019ICCV}. Most of these methods focus on learning a network mapping from different hand-crafted representations of face videos (e.g., cropped video frames~\cite{spetlik2018BMVC,yu2019ICCV}, motion representation~\cite{chen2018deepphys} or spatial-temporal map~\cite{niu2019rhythmnet}) to the physiological signals. One fact is that, these hand-crafted representations contain not only the information of physiological signals, but also the non-physiological information such as head movements, lighting variations and device noises. The features learned from these hand-crafted representations are usually affected by the non-physiological information.
In addition, existing methods use either the rPPG signals~\cite{chen2018deepphys,yu2019ICCV} or the average HR values~\cite{spetlik2018BMVC,niu2019rhythmnet} for supervision. However, rPPG signals and average HR describe the subject's physiological status from detailed and general aspects respectively, and making full use of both two physiological signals for training will help the network to dig more representative features.

To address these problems, we propose an end-to-end multi-task physiological measurement network, and train the network using a cross-verified disentangling (CVD) strategy. We first compress the face videos into multi-scale spatial-temporal maps (MSTmaps) to better represent the physiological information in face videos. Although the MSTmaps could highlight the physiological information in face videos, they are still polluted by the non-physiological information such as head movements, lighting conditions and device variations. To automatically disentangle the physiological features with the non-physiological information, we then use a cross-verified disentangling strategy to train our network.

As illustrated in Fig.~\ref{fig:overview}, this disentanglement is realized by designing an autoencoder with two encoders (one for physiological signals and the other for non-physiological information) and using pairwise MSTmaps as input for training. Besides of reconstructing the original MSTmaps, we also cross-generate the pseudo MSTmaps using the physiological and non-physiological features encoded from different original MSTmaps. Intuitively, the physiological and non-physiological features encoded from the pseudo MSTmap are supposed to be similar to the features used to generate the pseudo MSTmap. This principle can be used to guide the encoders to cross-verify the features they are supposed to encode, and thus make sure that physiological encoder only focuses on extracting features for HR and rPPG signal estimation, and the non-physiological encoder to encode other irrelevant information.
The physiological features are then used for multi-task physiological measurements, i.e., estimating average HR and rPPG signals synchronously. Results on multiple datasets for different physiological measurement tasks as well as the cross-database testing show the effectiveness of our method.

The contributions of this work include: 1) we propose a novel multi-scale spatial-temporal map to highlight the physiological information in face videos. 2) We propose a novel cross-verified disentangling strategy to distill the physiological features for robust physiological measurements. 3) We propose a multi-task physiological measurement network trained with our cross-verified disentangling strategy, and achieve the state-of-the-art performance on multiple physiological measurement databases as well as the cross-database testing.

\section{Related Work}

\subsection{Video-based Remote Physiological Measurement}

An early study of rPPG based physiological measurement was reported in~\cite{verkruysse2008remote}. After that, many approaches have been reported on this topic. The traditional hand-crafted methods mainly focus on extracting robust physiological signals using different color channels~\cite{poh2010non,poh2011advancements,li2014remote,lewandowska2011measuring} or different regions of interest (ROIs)~\cite{Lam2015Robust,Tulyakov2016Self}. It has been demonstrated that signal decomposition methods, such as independent component analysis (ICA)~\cite{poh2010non,poh2011advancements,Lam2015Robust}, principal components analysis (PCA)~\cite{lewandowska2011measuring} and matrix completion~\cite{Tulyakov2016Self}, are effective to improve the signal-to-noise rate (SNR) of the generated signal.
Besides the signal decomposition methods, there are also some approaches aimming to get a robust rPPG signal using reflection model based color space transformation~\cite{de2013robust,wang2015exploiting,wang2017amplitude}. Various hand-crafted methods using different signal decomposition methods and skin models lead to improved robustness. However, in less-constrained and complicated scenarios, the assumptions of the signal decomposition methods and skin reflection models may not hold, and the estimation performance will drop significantly.

In addition to these hand-crafted methods, there are also a few approaches aiming to leverage the effectiveness of deep learning to remote physiological measurement. In~\cite{chen2018deepphys}, Chen et al. proposed a motion representation of face video and used a convolution network with attention mechanism to predict the rPPG signals. In~\cite{niu2019rhythmnet}, Niu et al. utilized a spatial-temporal map as the representation of face video and used a CNN-RNN structure to regress the average HR value. In~\cite{spetlik2018BMVC}, Spetlik et al. proposed a full convolutional network to estimate the average HR from the cropped faces. In addition to these methods using 2D convolution, Yu et al.~\cite{yu2019ICCV} proposed a 3D convolution network to regress the rPPG signals from face videos. All these existing methods mainly focused on designing the effective representation of the input face videos~\cite{chen2018deepphys,niu2019rhythmnet} and using various network structures~\cite{spetlik2018BMVC,yu2019ICCV} for physiological measurement. However, non-physiological information such as head movements and lighting conditions may have significant impacts on these hand-crafted representations, but few of these methods have considered the non-physiological influences. In addition, these methods only use either rPPG signals or average HR values during training, and did not consider taking advantage of both physiological signals for supervision.

\subsection{Disentangle Representation Learning}

Disentanglement representation learning is gaining increasing attention in computer vision. In~\cite{luan2017disentangled}, Tran et al. proposed the DR-GAN to disentangle the identity and pose representations for pose-invariant face recognition with adversarial learning. In~\cite{yu2018exploring}, Liu et al. utilized an auto-encoder model to distill the identity and disentangle the identity features with other facial attributes. In~\cite{zhang2019gait}, Zhang et al. utilized the walking videos to disentangle the appearance and pose features for gait recognition. Besides the works on recognition tasks, there are also many works using the disentangled representation for image synthesis and editing. In~\cite{lu2019unsupervised}, Lu et al. proposed a method for image deblurring by splitting the content and blur features in a blurred image. In~\cite{Lee2019diverse}, Lee et al. proposed an approach for diverse image-to-image translation by embedding images into a domain-invariant content space and a domain specific attribute space.
Different from the existing approaches~\cite{luan2017disentangled,lu2019unsupervised,Lee2019diverse}, our method does not need adversarial training, making it more accessible for training. Besides, unlike~\cite{yu2018exploring}, which requires prior attribute labels for disentanglement, our method does not need the prior non-physiological information for training. In addition, unlike~\cite{zhang2019gait} requiring temporal information, our method only needs pairwise MSTmaps for disentanglement. Moreover, this work is the first work leveraging disentangle representation learning to remote physiological measurement.

\section{Proposed Method}

\begin{figure}[t]
      \centering
      \includegraphics[width=\linewidth]{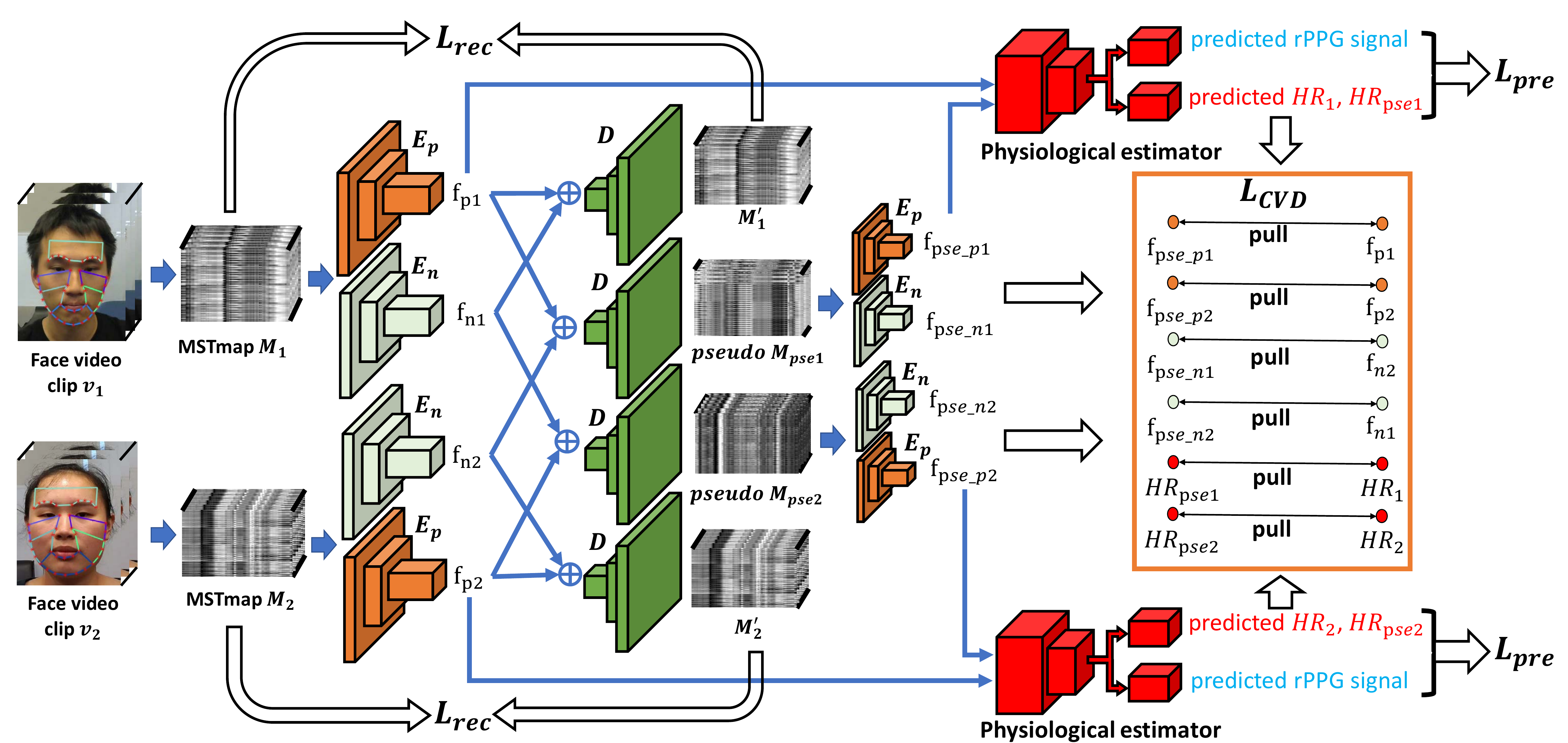}
      \caption{An overview of our cross-verified feature disentangling (CVD) strategy. Pairwise face video clips are used for training. We first generate the corresponding MSTmaps $M_1, M_2$ of the input face video clips. Then we feed the MSTmaps into the physiological encoder $E_p$ and non-physiological encoder $E_n$ to get the features. $L_{rec}$ is first used for reconstructing the original MSTmaps. Then we cross-generate the pseudo MSTmaps $M_{pse1}, M_{pse2}$ using features from different original MSTmaps. Disentanglement is realized using $L_{CVD}$ by cross-verifying the encoded features of the original MSTmaps $M_1, M_2$ and the pseudo MSTmaps $M_{pse1}, M_{pse2}$. The physiological estimator takes the physiological features $f_{p1}, f_{p2}, f_{pse\_p1}, f_{pse\_p2}$ as input and is optimized by $L_{pre}$. The modules of the same type in our network use shared weights.}
      \label{fig:overview}
\end{figure}

In this section, we give detailed explanations of the proposed cross-verified feature disentangling strategy for multi-task physiological measurement from face videos. Fig.~\ref{fig:overview} gives an overview of the proposed method, which includes the multi-scale spatial-temporal map generation, the cross-verified disentangling strategy and multi-task physiological measurements.


\subsection{Multi-scale Spatial Temporal Map}
\label{ms-STmap}

The amplitude of optical absorption variation of face skin is very small, thus it is important to design a good representation to highlight the physiological information in face videos~\cite{chen2018deepphys,niu2019rhythmnet}. Considering both the local and the global physiological information in face, we propose a novel multi-scale spatial-temporal map (MSTmap) to represent the facial skin color variations due to heartbeats.

\begin{figure}[t]
      \centering
      \includegraphics[width=0.8\linewidth]{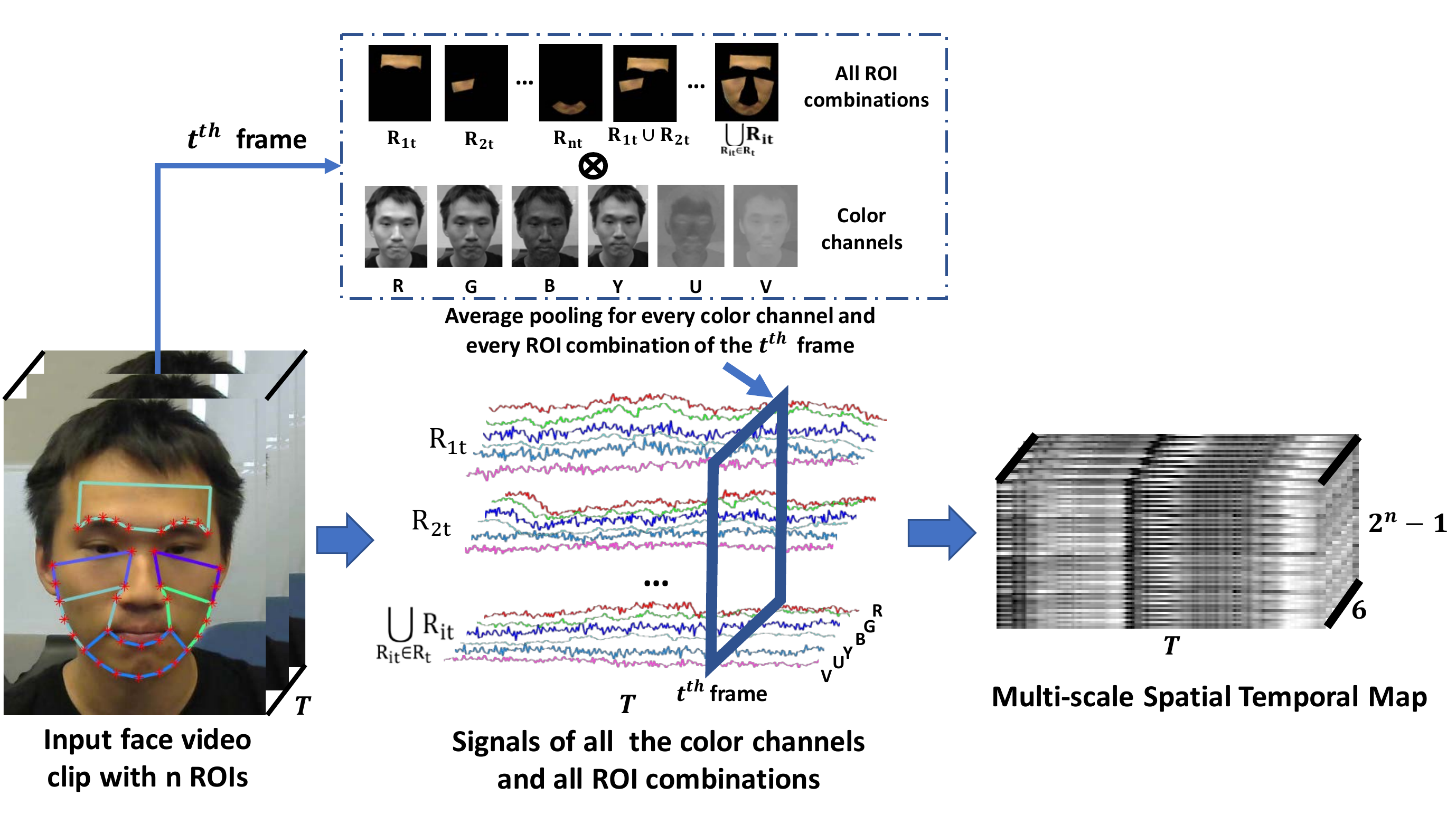}
      \caption{An illustration of our MSTmap generation from an input face video clip of $T$ frames. For the $t^{th}$ frame, we first detect the $n$ most informative ROIs in face, then calculate the average pixel values for all kinds of ROI combinations and all the color channels in both RGB and YUV color spaces. The average pixel values of different frames are concatenated into temporal sequences, and the temporal sequences of all ROI combinations and color channels are placed into rows. The final multi-scale spatial-temporal map is in the size of $(2^n-1)\times T \times 6$.}
      \label{fig:stmap}
\end{figure}

As shown in Fig.~\ref{fig:stmap}, we first use an open source face detector SeetaFace\footnote{\url{https://github.com/seetaface/SeetaFaceEngine}} to detect the facial landmarks, based on which we define the most informative ROIs for physiological measurement in face, i.e., the forehead and cheek areas. In order to stabilize the landmarks, we also apply a moving average filter to the facial landmark locations across frames. Average pooling is widely used to improve the robustness~\cite{poh2010non,poh2011advancements,li2014remote,lewandowska2011measuring,Tulyakov2016Self,wang2017algorithmic} against the background and device noises, and the traditional methods usually use the average pixel values of the whole facial ROI for further processing. Different from using the global pooling of the whole facial ROI, which may ignore some local information, we generate our MSTmap considering both the local and global facial regions.

Specifically, as shown in Fig.~\ref{fig:stmap}, for the $t^{th}$ frame of a video clip, we first get a set of $n$ informative regions of face $R_t = \{R_{1t}, R_{2t}, \cdots, R_{nt}\}$. Then, we calculate the average pixel values of each color channel for all the non-empty subsets of $R_t$, which are $2^n-1$ combinations of the elements in $R_t$. As illustrated in~\cite{niu2019rhythmnet}, YUV color space is effective in representing the physiological information in face. Therefore, we use both the RGB and YUV color space to calculate the MSTmap. The total number of color channels is $6$. For each video clip with $T$ frames, we first obtain $6\times (2^n-1)$ temporal sequences of length $T$. A max-min normalization is applied to all the temporal sequences to scale the temporal series into [0,255]. Then, the $2^n-1$ temporal signals of each channel are placed into rows, and we can get the final MSTmap with the size of $(2^n-1)\times T \times 6$ for each video clip.

\subsection{Cross-verified Feature Disentangling}

After we generate the MSTmaps as stated in Section~\ref{ms-STmap}, we can further use these informative representations for physiological measurements. One straight forward way is directly using convolution neural networks (CNNs) to regress the physiological signals with the MSTmaps. However, these hand-crafted MSTmaps are usually polluted by non-physiological information such as head movements and illumination conditions. In order to disentangle the physiological feature from the non-physiological information, we use a cross-verified disentangling strategy (CVD) to train the network. An auto-encoder with two encoders (one for physiological features and the other for non-physiological information) is used to learn the features. Pairwise MSTmaps are used as input, and the network is trained with both the original MSTmaps as well as the pseudo MSTmaps cross-decoded using features from different original MSTmaps.

Specifically, as shown in Fig.~\ref{fig:overview}, with pairwise input face video clips $v_1, v_2$, we first generate the corresponding MSTmaps $M_1, M_2$. Then, we use the physiological encoder $E_p$ and noise encoder $E_n$ to get the physiological features $f_{p1}, f_{p2}$ and the non-physiological features $f_{n1}, f_{n2}$. Physiological and non-physiological features encoded from the same MSTmap are first used to reconstruct the original MSTmap to make sure the decoder $D$ can effectively reconstruct the MSTmap. For the input MSTmaps $M_1, M_2$ and the decoded MSTmaps $M_1^{'},M_2^{'}$, the reconstruction loss is formulated as
\begin{equation}
L_{rec} = \lambda_{rec} \sum_{i=1}^{2} \parallel M_i - M_i^{'}\parallel_1
\end{equation}
where $\lambda_{rec}$ is used to balance $L_{rec}$ and other losses.

Then, pseudo MSTmaps $M_{pse1}, M_{pse2}$ are generated with the decoder $D$ using the features from different MSTmaps, i.e., pseudo MSTmap $M_{pse1}$ is generated using $f_{p1}$ and $f_{n2}$, and $M_{pse2}$ is generated using $f_{p2}$ and $f_{n1}$. $M_{pse1}, M_{pse2}$ are further fed to $E_p$ and $E_n$ to get the encoded features $f_{pse\_p1}, f_{pse\_n1}, f_{pse\_p2}, f_{pse\_n2}$. Meanwhile, all the physiological features $f_{p1}$, $f_{p2}$, $f_{pse\_p1}$ and $f_{pse\_p2}$ are fed to the physiological estimator for HR and rPPG signal predictions. The detailed architecture of $E_p$, $E_n$ and $D$ can be found in the supplementary material.

Institutively, taking $f_{p1}$ and $f_{pse\_p1}$ as example, if the physiological encoder $E_p$ encodes only the physiological information, $f_{p1}$ is supposed to be the same as $f_{pse\_p1}$ since $M_{pse1}$ are generated using $f_{p1}$ and $f_{n2}$ and the physiological representation contained in $M_{pse1}$ should be the same as $M_1$. This principle can be used to cross-verify the features of the two encoders, and guide the two encoders to focus on the information they are supposed to encode. We consider the physiological features, non-physiological features as well as the predicted HR values $HR_1, HR_{pse1}, HR_2, HR_{pse2}$ with the physiological features, and design our CVD loss $L_{CVD}$ as:
\begin{equation}
\begin{aligned}
L_{CVD} &= \lambda_{cvd}\sum_{i=1}^{2} \parallel f_{pi} - f_{pse\_pi}\parallel_1 + \lambda_{cvd} \sum_{i=1}^{2} \parallel f_{ni} - f_{pse\_n(3-i)}\parallel_1 \\
&+ \sum_{i=1}^{2} \parallel HR_{i} - HR_{psei}\parallel_1 \\
\end{aligned}
\end{equation}
where $\lambda_{cvd}$ is the balance parameter.
With this CVD loss, we can enforce the physiological and non-physiological encoders to obtain features disentangled from each other, and make the physiological encoder to get more representative features for physiological measurements.


\subsection{Multi-task Physiological Measurement}

The physiological features are further used for multi-task physiological measurements. Typically-considered physiological signals include the average HR values and the rPPG signals. The average HR values can provide general information of the physiological status, while the rPPG signals can give more detailed supervision. Considering that these two signals can provide different aspects of supervision for training, we design our physiological estimator taking both of them into consideration, and expect that this will help the network to learn more robust features and get more accurate predictions.

Specifically, our physiological estimator is a two-head network with one head to regress the rPPG signals and the other to regress the HR values. The detailed architectures can be found in the supplementary material. For the average HR value regression branch, we use a conventional L1 loss function for supervision. For the rPPG signal prediction branch, we use a Pearson correlation based loss to define the similarity between the predicted signal and ground truth, i.e.,
\begin{equation}
L_{rppg} = 1 - \frac{Cov(s_{pre}, s_{gt})}{\sqrt{Cov(s_{pre}, s_{pre})}\sqrt{Cov(s_{gt}, s_{gt})}}
\end{equation}
where $s_{pre}$ and $s_{gt}$ are the predicted and ground truth rPPG signals, and $Cov(x,y)$ is the covariance of $x$ and $y$. Meanwhile, since average HR can also be calculated from the rPPG signal, we add the $L_{rppg\_hr}$ to help the average HR estimation using the features of the rPPG branch. The $L_{rppg\_hr}$ is formulated as
\begin{equation}
L_{rppg\_hr} = CE(PSD(s_{pre}), HR_{gt})
\end{equation}
where $HR_{gt}$ is the ground-truth HR, and $CE(x, y)$ calculate the cross-entropy loss of the input $x$ and ground-truth $y$. $PSD(s)$ is the power spectral density of the input signal $s$. The final loss for physiological measurements is
\begin{equation}
L_{pre} = \|HR_{pre} - HR_{gt}\| + \lambda_{rppg} L_{rppg} + L_{rppg\_hr}
\end{equation}
where $HR_{pre}$ and $HR_{gt}$ are the predicted HR and ground-truth HR, and $\lambda_{rppg}$ is a balancing parameter. The overall loss function of our cross-verified disentangling strategy is
\begin{equation}
L = L_{rec} + L_{CVD} + L_{pre}
\end{equation}

\section{Experiments}
In this section, we provide evaluations of the proposed method including intra-database testing, cross-database testing and the ablation study.

\subsection{Databases and Experimental Settings}
\subsubsection{Databases}

We evaluate our method on three widely-used publicly-available physiological measurement databases, i.e., VIPL-HR, OBF, and MMSE-HR.

 \textbf{VIPL-HR}~\cite{niu2019rhythmnet,niu2018VIPL-HR} is a large-scale remote HR estimation database containing 2,378 visible light face videos from 107 subjects. Various less-constrained scenarios, including different head movements, lighting conditions and acquisition devices, are considered in this database. The frame rates of the videos in VIPL-HR vary from 25 fps to 30 fps. Average HR values as well as the BVP signals are provided, and we use the BVP signals as the ground truth rPPG signals. Following the protocol in~\cite{niu2019rhythmnet}, we conduct a five-fold subject-exclusive cross-validation for the intra-database testing as well as the ablation study for average HR estimation.

 \textbf{OBF}~\cite{li2018obf} is a large-scale database for remote physiological signal analysis. It contains 200 five-minute-long high-quality RGB face videos from 100 subjects. The videos were recorded at 60 fps in OBF, we downsample the frame rate to 30 fps for the convenience of computing. Following~\cite{yu2019ICCV}, we use OBF for evaluations on both HR estimation, RF measuring and HRV analysis. The ground truth RF and HRV features are calculated using the corresponding ECG signals provided. The BVP signals provided are used as the ground truth rPPG signals. We conduct a ten-fold subjective-exclusive cross-validation as~\cite{yu2019ICCV}.

 \textbf{MMSE-HR}~\cite{Tulyakov2016Self} is a database for remote HR estimation consisting of 102 RGB face videos from 40 subjects recorded at 25 fps. The corresponding average HR values are collected using a biological data acquisition system. Various facial expressions and head movements of the subjects are recorded. The MMSE-HR database is only used for cross-database testing.

\subsubsection{Training Details}

The detailed architectures of the network can be found in the supplementary material, and all the losses are applied jointly. For all the experiments, the length of face video clip is set to 300 frames, and 6 ROIs as shown in Fig.~\ref{fig:stmap} are considered. For average HR estimation of a 30-second face video as~\cite{niu2019rhythmnet}, we use a time step of 0.5s to get all the video clips, and the average of the predicted HRs is regarded as the predicted average HR for the 30-second video. The MSTmaps are resized to $320\times 320$ before being fed to the network for the convenience of computing. Random horizontal and vertical flip of the MSTmaps as well as the data balancing strategy proposed in~\cite{niu2019robust} are used for data augmentation. For all the experiments, we set $\lambda_{rec}$ = 50, $\lambda_{cvd}$ = 10 and $\lambda_{rppg}$ = 2.
All the networks are implemented using PyTorch framework\footnote{\url{https://pytorch.org/}} and trained with NVIDIA P100. Adam optimizer~\cite{kingma2014adam} with an initial learning rate of 0.0005 is used for training. The maximum epoch number for training is set to 70 for experiments on VIPL-HR database and 30 for experiments on OBF database. Code is available.\footnote{\url{https://github.com/nxsEdson/CVD-Physiological-Measurement}}

\subsubsection{Evaluation Metrics}

Various metrics are used for evaluations. For the task of average HR estimation, we follow previous work~\cite{niu2019rhythmnet,Tulyakov2016Self,li2014remote,yu2019ICCV} and use the metrics including the standard deviation of the error (Std), the mean absolute error (MAE), the root mean squared error (RMSE), and the Pearson's correlation coefficients ($r$). For the evaluation of RF and HRV analysis, following~\cite{li2018obf,yu2019ICCV}, we use Std, RMSE and $r$ as the evaluation of RF and three HRV features, i.e., low frequency (LF), high frequency (HF) and LF/HF.

\subsection{Intra-database Testing}
\subsubsection{Results on Average HR Estimation}

We first conduct experiments on VIPL-HR database for average HR estimation using a five-fold subject-exclusive evaluation protocol following~\cite{niu2019rhythmnet}. State-of-the-art methods including hand-crafted methods (SAMC~\cite{Tulyakov2016Self}, POS~\cite{wang2017algorithmic}, CHROM~\cite{de2013robust}) and deep learning based methods (I3D~\cite{carreira2017quo}, DeepPhy~\cite{chen2018deepphys}, RhythmNet~\cite{niu2019rhythmnet}) are used for comparison. We directly take the results of these state-of-the-art methods from~\cite{niu2019rhythmnet}. The results of the proposed method and the state-of-the-art methods are given in Table~\ref{table:vipl_color}.

\setlength{\tabcolsep}{5pt}
\begin{table}[t]
\begin{center}
\caption{The HR estimation results by the proposed approach and several state-of-the-art methods on the VIPL-HR database.}
\label{table:vipl_color}
\begin{tabular}{|l|cccc|}
\hline
\multirow{2}{*}{Method} & Std$\downarrow$  & MAE$\downarrow$ & RMSE$\downarrow$  & \multirow{2}{*}{$r\uparrow$}\\
&(bpm) &(bpm) &(bpm)&\\
\hline
\hline
SAMC~\cite{Tulyakov2016Self} & 18.0 & 15.9 & 21.0 & 0.11\\
POS~\cite{wang2017algorithmic} & 15.3 & 11.5 & 17.2 & 0.30\\
CHROM~\cite{de2013robust} & 15.1 & 11.4 & 16.9 & 0.28 \\
I3D~\cite{carreira2017quo} & 15.9 & 12.0 & 15.9 & 0.07\\
DeepPhy~\cite{chen2018deepphys} & 13.6 & 11.0 & 13.8 &  0.11\\
RhythmNet~\cite{niu2019rhythmnet} & 8.11 & 5.30 & 8.14  & 0.76\\
\hline
Proposed & \textbf{7.92} & \textbf{5.02} & \textbf{7.97} &\textbf{0.79} \\
\hline
\end{tabular}
\end{center}
\end{table}

\begin{figure}[t]
\centering
\subfigure[]{
\includegraphics[width=0.3\linewidth]{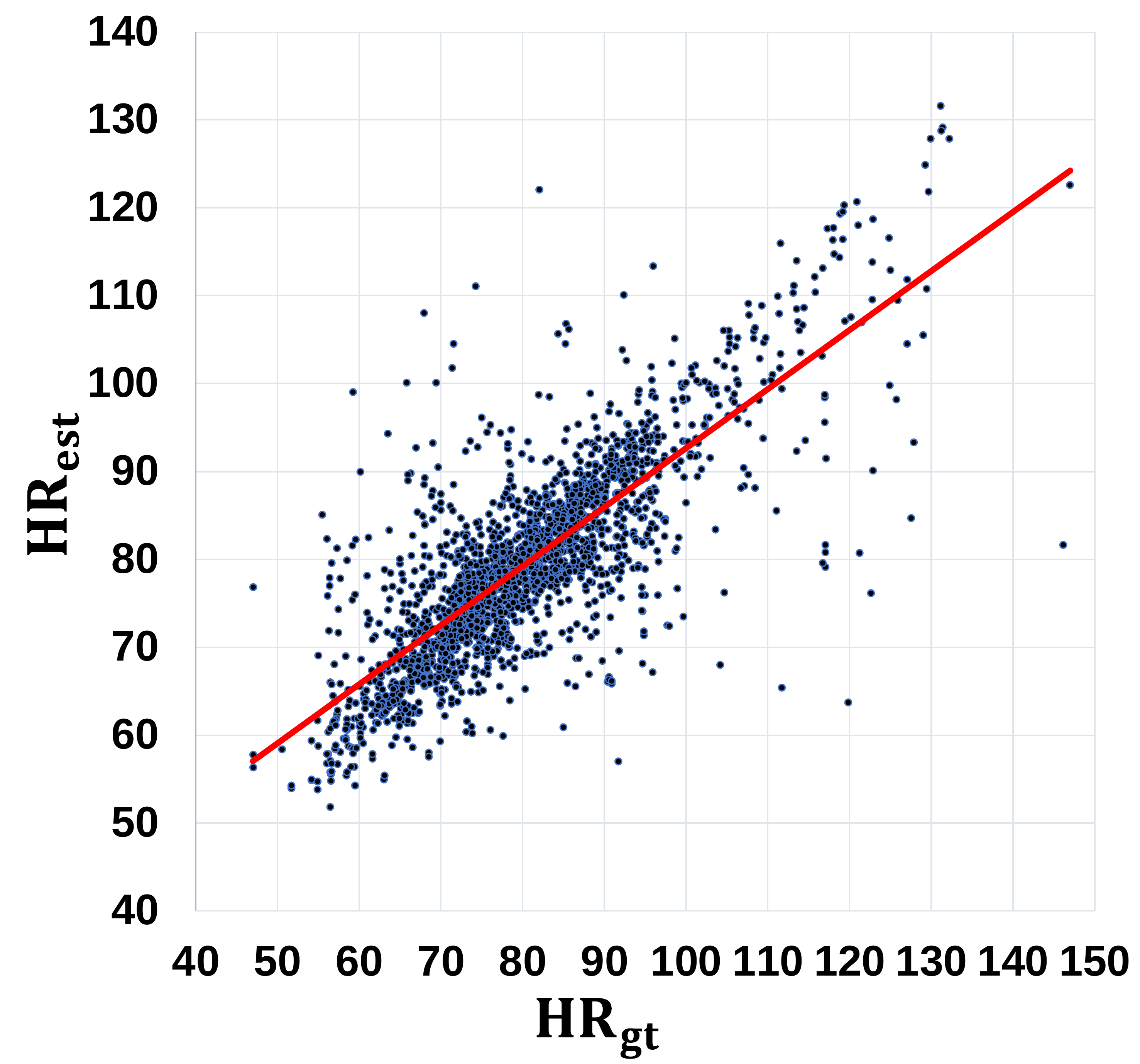}
\label{fig:HR_plot}
}
\subfigure[]{
\includegraphics[width=0.3\linewidth]{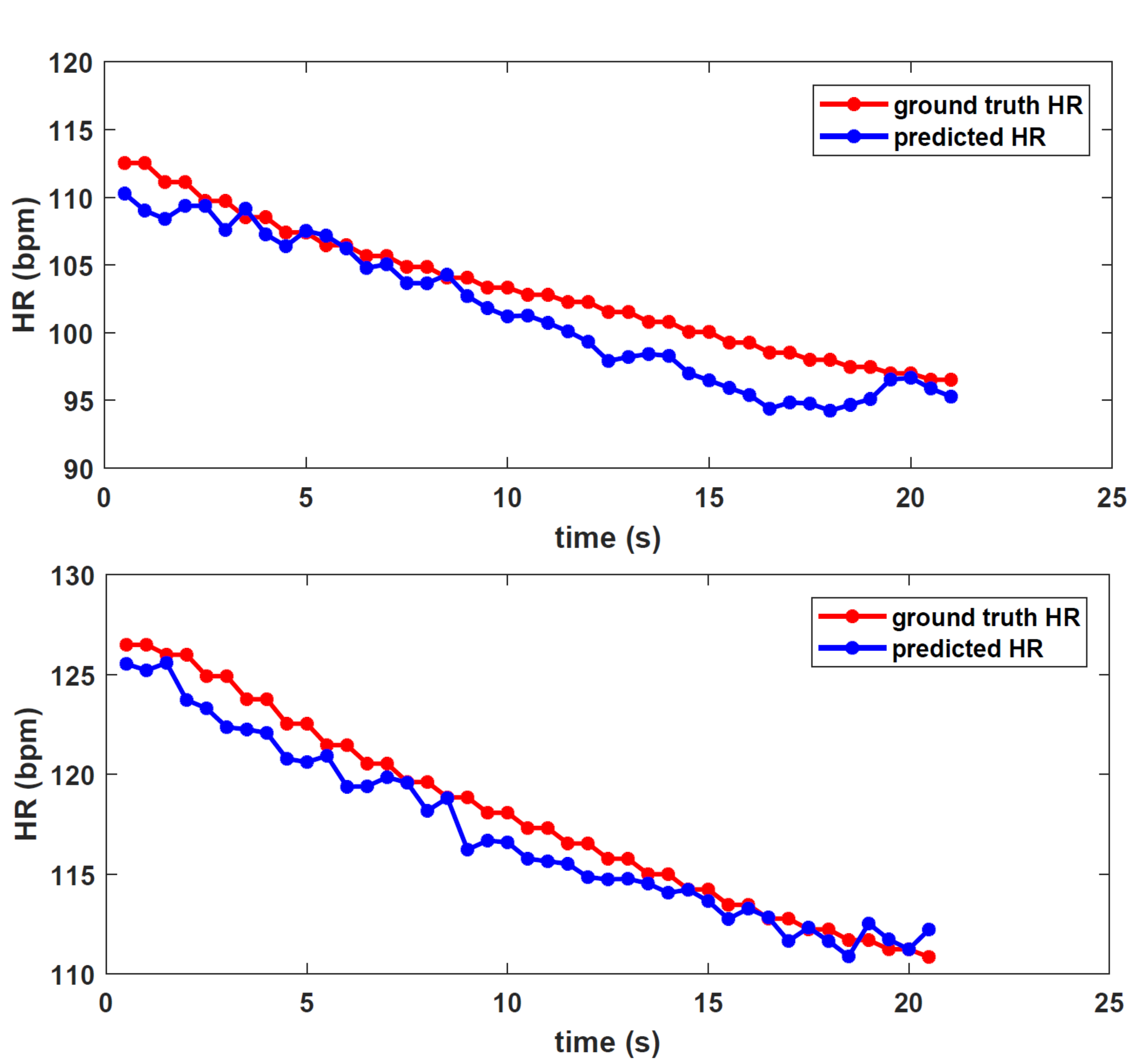}
\label{fig:HR_decrease}
}
\subfigure[]{
\includegraphics[width=0.3\linewidth]{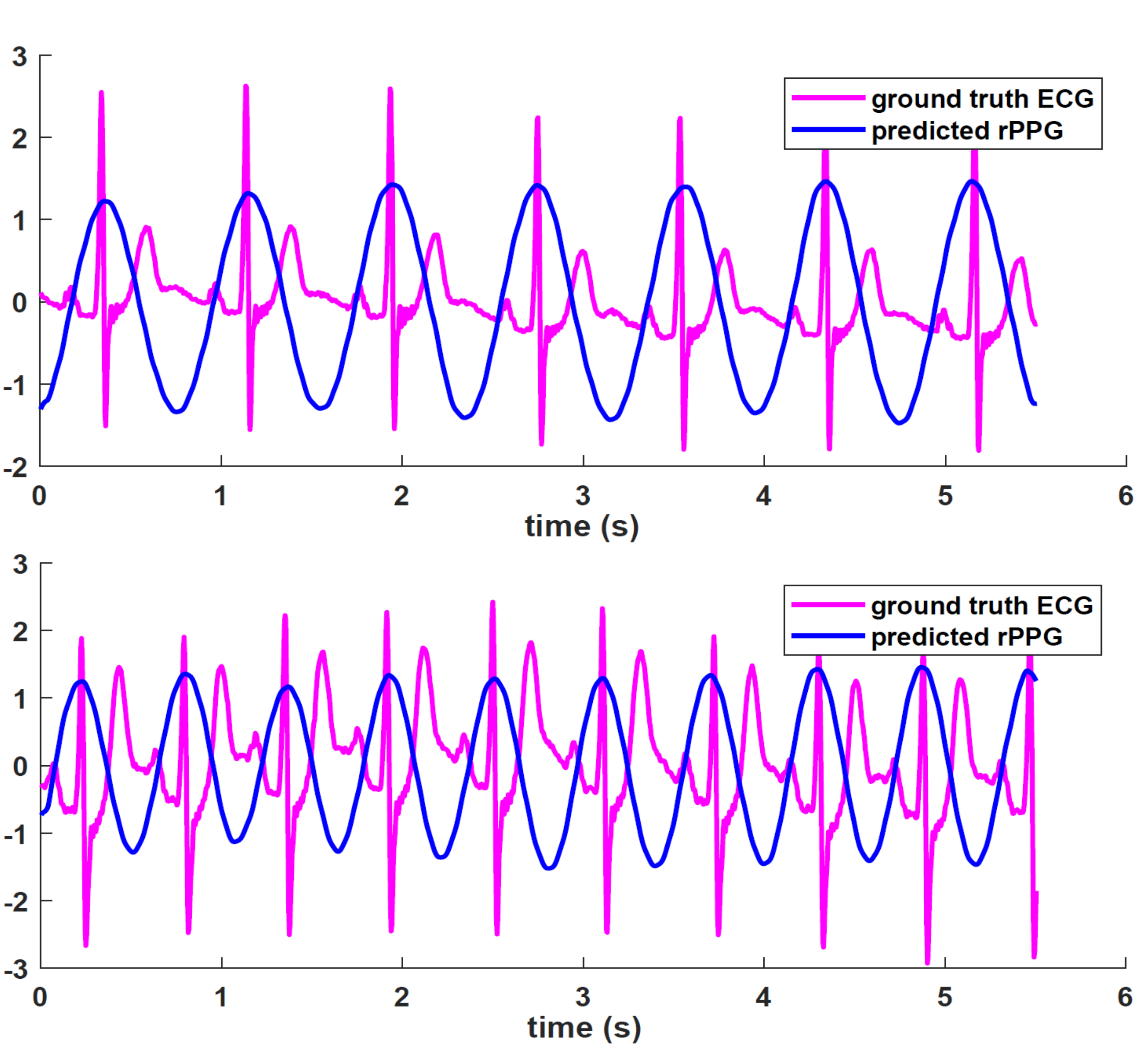}
\label{fig:HR_curve}
}
\caption{(a) The scatter plot of the ground truth $HR_{gt}$ and the predicted $HR_{est}$ of all the face videos on VIPL-HR dataset. (b) Two examples of the short-time HR estimation for face videos with significantly decreased HR. (c) Two example curves of the predicted rPPG signals and the ground truth ECG signals used to calculate the HRV features.}
\label{fig:VIPL_analysis}
\end{figure}

From the results, we can see that the proposed method achieves promising results with an Std of 7.92 bpm, an MAE of 5.02 bpm, an RMSE of 7.97 bpm and a $r$ of 0.79, which outperform all the state-of-the-art methods including both the hand-crafted and deep learning based methods. In order to further check the correlations between the predicted HRs and the ground-truth HRs, we plot the HR estimation results against the ground truths in Fig.~\ref{fig:HR_plot}. From the figure we can see that the predicted HRs and the ground-truth HRs are well correlated in a wide range of HR from 47 bpm to 147 bpm under the less-constrained scenarios of the VIPL-HR database such as large head movements and dim environment. In addition, we also calculate the estimation errors for the large head movement scenario in VIPL-HR and compare the result with RhythmNet~\cite{niu2019rhythmnet}. We get an RMSE of 7.44bpm, which is distinctively better than RhythmNet (9.4bpm). All the results indicate that the proposed method could effectively distill the physiological information and provide robust physiological measurements.

Besides the average HR estimation for a thirty-second video, we also check the short-time HR estimation performance of the after exercising scenario on the VIPL-HR, in which the subject's HR decreases rapidly. Two examples are given in Fig.~\ref{fig:HR_decrease}. From the examples, we can see that the proposed approach could follow the trend of HR changes well, which indicates our model is robust in the significant HR changing scenarios.

In addition to experiments on VIPL-HR, we also evaluate the average HR estimation performance on OBF. State-of-the-art methods including ROI$_{green}$~\cite{li2018obf}, CHROM~\cite{de2013robust}, POS~\cite{wang2017algorithmic} and rPPGNet~\cite{yu2019ICCV} are used for comparison, and the results of these methods are from~\cite{yu2019ICCV}. As shown in Table~\ref{table:obf_hr}, our method also achieves the best performance, indicating the effectiveness of the proposed approach.

\subsubsection{Results on RF Measurement and HRV Analysis}

We also conduct experiments for RF measurement and HRV analysis on the OBF database. Following~\cite{yu2019ICCV}, we use a 10-fold subject-exclusive protocol for all experiments. RF and three HRV features, i.e., low frequency (LF), high frequency (HF) and LF/HF are considered for evaluations. We compare the proposed method with the state-of-the-art methods including hand-crafted methods ROI$_{green}$~\cite{li2018obf}, CHROM~\cite{de2013robust} and POS~\cite{wang2017algorithmic} and the learning based method rPPGNet~\cite{yu2019ICCV}. The results of ROI$_{green}$~\cite{li2018obf}, CHROM~\cite{de2013robust}, POS~\cite{wang2017algorithmic} and rPPGNet~\cite{yu2019ICCV} are taken from~\cite{yu2019ICCV}. All the results are shown in Table~\ref{table:obf}.

From the results, we can see that the proposed approach outperforms all the existing state-of-the-art methods by a large margin on all evaluation metrics for RF and all HRV features. These results indicate that our method could not only handle the average HR estimation task but also could give a promising prediction of the rPPG signal for RF measurement and HRV analysis, which show the effectiveness of the proposed method and the potential in many healthcare applications. We further check the predicted rPPG signals of our estimator. Two examples are given in Fig.~\ref{fig:HR_curve}. From the results we can see that our method could give an accurate prediction of the interbeat intervals (IBIs), thus can give a robust estimation of RF and HRV features.

\setlength{\tabcolsep}{0.3pt}
\begin{table}[t]
\begin{center}
\caption{The results of RF measurement and HRV analysis by the proposed approach and several state-of-the-art methods on the OBF database.}
\vspace{-0.4cm}
\label{table:obf}
\begin{tabular}{|l|ccc|ccc|ccc|ccc|ccc|}
\noalign{\smallskip}
\hline
\multirow{2}{*}{Method} & \multicolumn{3}{|c}{RF(Hz)} & \multicolumn{3}{|c}{LF(u.n)} & \multicolumn{3}{|c}{HF(u.n)} & \multicolumn{3}{|c|}{LF/HF}\\
\cline{2-13}
& Std & RMSE & $r$ & Std & RMSE & $r$ & Std & RMSE & $r$ & Std & RMSE & $r$\\
\hline
\hline
ROI$_{green}$~\cite{li2018obf} & 0.078 & 0.084 & 0.321 & 0.22 & 0.24 & 0.573 & 0.22 & 0.24 & 0.573 & 0.819 & 0.832 & 0.571\\
CHROM~\cite{de2013robust} & 0.081& 0.081& 0.224& 0.199& 0.206& 0.524& 0.199& 0.206& 0.524& 0.83& 0.863& 0.459\\
POS~\cite{wang2017algorithmic} &0.07 &0.07 &0.44 &0.155 &0.158 &0.727 &0.155 &0.158 &0.727 &0.663& 0.679& 0.687\\
rPPGNet\cite{yu2019ICCV} & 0.064& 0.064& 0.53& 0.133& 0.135& 0.804& 0.133& 0.135& 0.804& 0.58& 0.589& 0.773\\
\hline
Proposed &\textbf{0.058} &\textbf{0.058} &\textbf{0.606} &\textbf{0.09} &\textbf{0.09} &\textbf{0.914} &\textbf{0.09} &\textbf{0.09} &\textbf{0.914} &\textbf{0.453} &\textbf{0.453} &\textbf{0.877}\\
\hline
\end{tabular}
\end{center}
\end{table}

\subsection{Cross-database Testing}

Besides of the intra-database testings on the VIPL-HR and OBF databases, we also conduct cross-database testing on the small-scale HR estimation database MMSE-HR following the protocol of~\cite{niu2019rhythmnet}. The model is trained on the VIPL-HR database and directly tested on the MMSE-HR.
State-of-the-art method including hand-crafted methods Li2014~\cite{li2014remote}, CHROM~\cite{de2013robust}, SAMC~\cite{Tulyakov2016Self} and deeply learned method RhythmNet~\cite{niu2019rhythmnet} are listed for comparison. The results of Li2014~\cite{li2014remote}, CHROM~\cite{de2013robust}, SAMC~\cite{Tulyakov2016Self} and RhythmNet~\cite{niu2019rhythmnet} are from~\cite{niu2019rhythmnet}. All the results of the proposed approach and the state-of-the-art methods are shown in Table~\ref{table:mmsehr}.



From the results, we can see that our model gets the best results on all evaluation metrics compared with the state-of-the-art methods. These results indicate that our method could help the network to have a good generalization ability to the unknown scenarios without any prior knowledge, which demonstrates the effectiveness of the proposed approach.

\setlength{\tabcolsep}{0.5pt}
\begin{minipage}[b]{\textwidth}
\begin{minipage}[b]{0.45\textwidth}
\centering
\makeatletter\def\@captype{table}\makeatother
\caption{The average HR estimation results by the proposed approach and several state-of-the-art methods on the OBF database.}
\label{table:obf_hr}
\smallskip
\begin{tabular}{|l|ccc|}
\hline
\multirow{2}{*}{Method} & Std$\downarrow$ & RMSE$\downarrow$ & \multirow{2}{*}{$r\uparrow$}\\
&(bpm) &(bpm) &\\
\hline
\hline
ROI$_{green}$~\cite{li2018obf} & 2.159 & 2.162 & 0.99 \\
CHROM~\cite{de2013robust} & 2.73 & 2.733& 0.98\\
POS~\cite{wang2017algorithmic} &1.899 &1.906 &0.991 \\
rPPGNet\cite{yu2019ICCV} & 1.758 & 1.8 & 0.992\\
\hline
Proposed & \textbf{1.257} & \textbf{1.26}  & \textbf{0.996}\\
\hline
\end{tabular}
\end{minipage}\hspace{15pt}
\begin{minipage}[b]{0.45\textwidth}
\centering
\makeatletter\def\@captype{table}\makeatother
\caption{The cross-database HR estimation results by the proposed approach and several state-of-the-art methods on the MMSE-HR database.}
\smallskip
\label{table:mmsehr}
\begin{tabular}{|l|ccc|}
\hline
\multirow{2}{*}{Method} & Std$\downarrow$ & RMSE$\downarrow$ & \multirow{2}{*}{$r\uparrow$}\\
&(bpm) &(bpm) &\\
\hline
\hline
Li2014~\cite{li2014remote} & 20.02 & 19.95 & 0.38 \\
CHROM~\cite{de2013robust} & 14.08 & 13.97  & 0.55 \\
SAMC~\cite{Tulyakov2016Self} &12.24 & 11.37 & 0.71 \\
RhythmNet~\cite{niu2019rhythmnet} & 6.98 & 7.33  & 0.78\\
\hline
Proposed & \textbf{6.06} & \textbf{6.04}  & \textbf{0.84}\\
\hline
\end{tabular}
\end{minipage}
\end{minipage}

%

\subsection{Ablation Study}

We also provide the results of ablation studies for the proposed method for HR estimation on the VIPL-HR database. All the results are shown in Table~\ref{table:ablation}.


\setlength{\tabcolsep}{4pt}
\begin{table}[t]
\begin{center}
\caption{The HR estimation results of the ablation study on the VIPL-HR database.}
\label{table:ablation}
\begin{tabular}{|l|cccc|}
\hline
\multirow{2}{*}{Method} & Std$\downarrow$  & MAE$\downarrow$ & RMSE$\downarrow$  & \multirow{2}{*}{$r\uparrow$}\\
&(bpm) &(bpm) &(bpm)&\\
\hline
\hline
MSTmap+HR &10.16 & 6.39 &10.24	&0.662\\
MSTmap+MTL & 8.93 & 5.55 & 9.03	&0.736\\
\hline
STmap+MTL &8.98 &5.80 &9.01 &0.727 \\
STmap+MTL+CVD &8.41 &5.34 &8.51 &0.765\\
\hline
MSTmap+MTL+$N_{movement}$ &8.81 &5.58 &8.96 &0.743 \\
MSTmap+MTL+$N_{stdface}$ &8.60 &5.46 &8.72 &0.756 \\
\hline
MSTmap+MTL+CVD(proposed) & \textbf{7.92} & \textbf{5.02} & \textbf{7.97} & \textbf{0.796} \\
\hline
\end{tabular}
\small{\text{\textbf{MTL}: multi-task learning; \textbf{CVD}: cross-verified disentangling}}
\end{center}
\vspace{-0.5cm}
\end{table}

\subsubsection{Effectiveness of Multi-task Learning}

In order to validate the effectiveness of the multi-task learning, we train our network using just the HR estimation branch (\emph{MSTmap+HR}) as well as using both HR and rPPG estimation branches (\emph{MSTmap+MTL}). The input of the network is our MSTmaps, and the network is trained without the cross-verified disentangling strategy. From the results, we can see that when we use both the rPPG and HR branches for training, the HR estimation results is better than only using the HR estimation branch. The MAE is reduced from 6.39 bpm to 5.55 bpm and the RMSE is reduced from 10.24 bpm to 9.03 bpm. These results indicate that rPPG signals and HR describe different aspects of the subject's physiological status, and learning from both rPPG signals and average HR values can help the network to learn both the general and the detailed features of physiological information, and thus benefit the average HR estimation.

\subsubsection{Effectiveness of Multi-scale Spatial Temporal Map}

We then test the effectiveness of our MSTmap. We use another effective physiological representation proposed by~\cite{niu2019rhythmnet} (STmap) for comparison. Experiments with and without the cross-verified disentangling strategy are conducted. When we train the network without the cross-verified disentangling strategy, using MSTmap as the physiological representation of face video (\emph{MSTmap+MTL}) outperforms using STmap (\emph{STmap+MTL}) as the representation with a comparable RMSE as well as a better MAE of 5.55 bpm and a better Pearson correlation coefficient of 0.736. When we apply the proposed cross-verified disentangling strategy during training, we can see that the model using our MSTmap (\emph{MSTmap+MTL+CVD}) outperforms the model using the STmap (\emph{STmap+MTL+CVD}) for all evaluations. These results indicate that our MSTmap is effective to represent the physiological information in face videos.

\subsubsection{Effectiveness of Cross-verified Disentangling}

We further evaluate the effectiveness of our CVD strategy. We first compare the results with and without using the CVD strategy during training. From the results, we can see that no matter what representation of face video we use as input, our CVD strategy could bring a large improvement to the final results. When using the MSTmap as input, training with cross-verified disentangling strategy (\emph{MSTmap+MTL+CVD}) can reduce the estimation RMSE from 9.03 bpm to 7.97 bpm and the MAE from 5.55 bpm to 5.02 bpm. When we use STmap as the representation (\emph{STmap+MTL+CVD}), our CVD strategy can again bring an improvement of the HR estimation accuracy. These results indicate that our cross-verified disentangling strategy could effectively improve the physiological measurement accuracy.

Besides the experiments with and without the cross-verified disentangling strategy, we also compare the proposed CVD strategy with training the network using pre-defined non-physiological signals for disentanglement. The disentanglement using pre-defined non-physiological signals is implemented by using a decoder with the same architecture as $D$, and decoded the non-physiological signals with $f_n$. In~\cite{wang2019Discriminative}, two pre-defined non-physiological signals, i.e., the head movements ($N_{movement}$) and the standard deviation of the facial pixel values ($N_{stdface}$), are used to improve the HR estimation accuracy. Following~\cite{wang2019Discriminative}, we also use these two non-physiological signals for disentanglement. The results are denoted as \emph{MSTmap+MTL+$N_{movement}$} and \emph{MSTmap+MTL+$N_{stdface}$}.

On one hand, we can see that both pre-defined non-physiological signals could help to reduce the HR estimation errors. When the head movement ($N_{movement}$) is used as the non-physiological signal, the disentanglement achieves a comparable MAE and slightly improves the Std, RMSE and $r$. When we use the standard deviation of the facial pixel values ($N_{stdface}$) as the non-physiological signals, it achieves better results on all evaluations because $N_{stdface}$ contains more non-physiological information of the face video. These results indicate that the MSTmaps are usually polluted by the non-physiological information, and disentangling strategy is necessary.
On the other hand, pre-defined non-physiological signals are only a subset of the non-physiological information of the MSTmaps. Our cross-verified disentangling strategy can help the network to distill the physiological features from data, which are more representative than using the pre-defined non-physiological signals. The results of using the cross-verified disentangling strategy outperform all the disentangling methods using pre-defined physiological signals on all evaluation metrics.
All these experiments indicate that our CVD strategy is effective to distill the physiological information and thus benefit the physiological measurements.

\section{Conclusions}

In this paper, we propose an effective end-to-end multi-task network for multiple physiological measurements using cross-verified disentangling strategy to reduce the influences of non-physiological signals such as head movements, lighting conditions, etc. The input face videos are first compressed into a hand-crafted representation named multi-scale spatial-temporal map to better represent the physiological information in face videos. Then we take pairwise MSTmaps as input and train the network with a cross-verified disentangling strategy to get effective physiological features. The learned physiological features are used for both the average HR estimation and rPPG signals regression. The proposed method achieves state-of-the-art performance in multiple physiological measurement tasks and databases. In our future work, we would like to explore the semi-supervised learning technologies for remote physiological measurement.


\textbf{Acknowledgment}
This work is partially supported by National Key R\&D Program of China (grant 2018AAA0102501), Natural Science Foundation of China (grant 61672496), the Academy of Finland for project MiGA (grant 316765), project 6+E (grant 323287), ICT 2023 project (grant 328115), and Infotech Oulu.

\section*{Supplementary Material}
\textbf{Detailed network architectures}

\begin{figure}
\centering
\subfigure[]{
\includegraphics[width=0.27\linewidth]{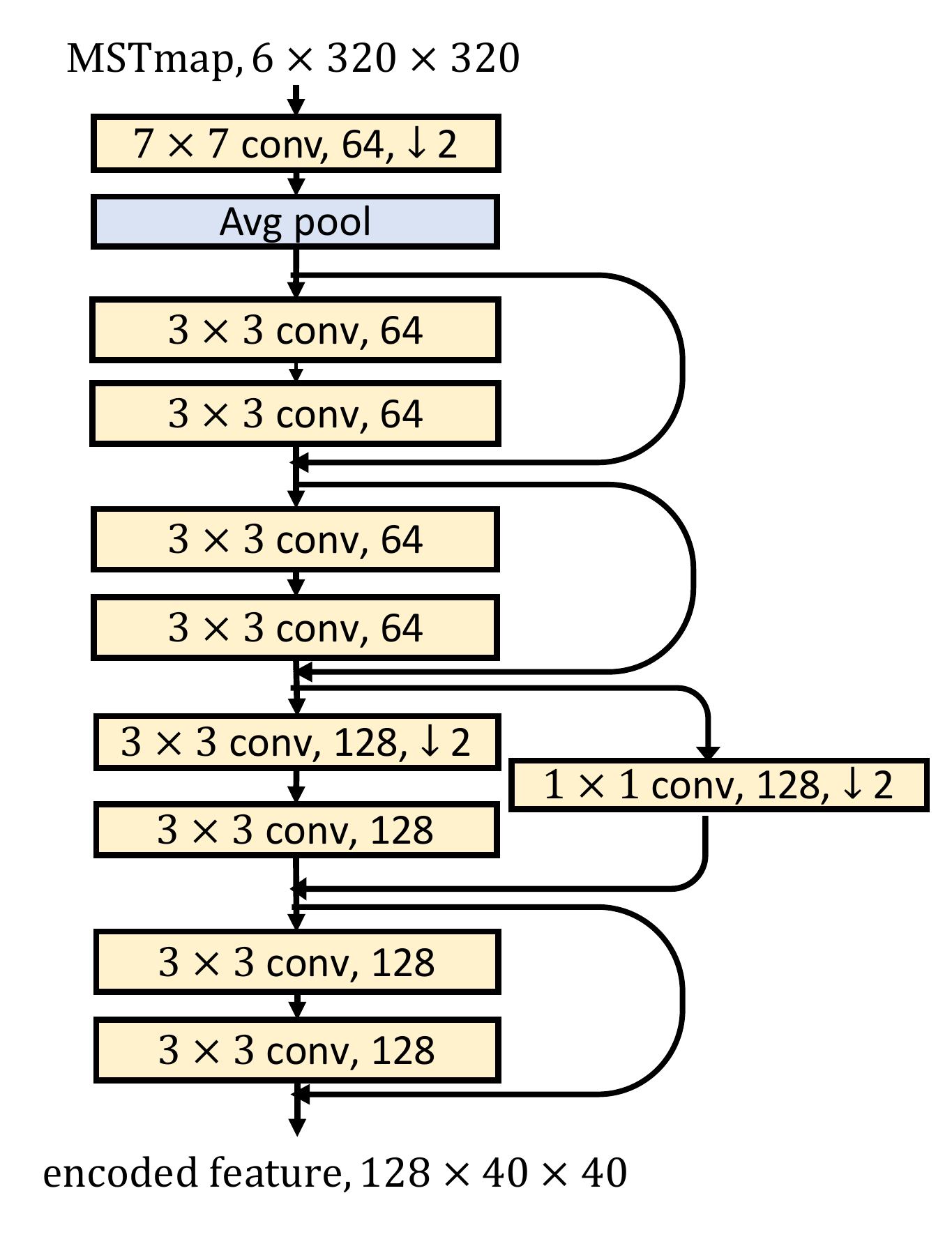}
}
\subfigure[]{
\includegraphics[width=0.27\linewidth]{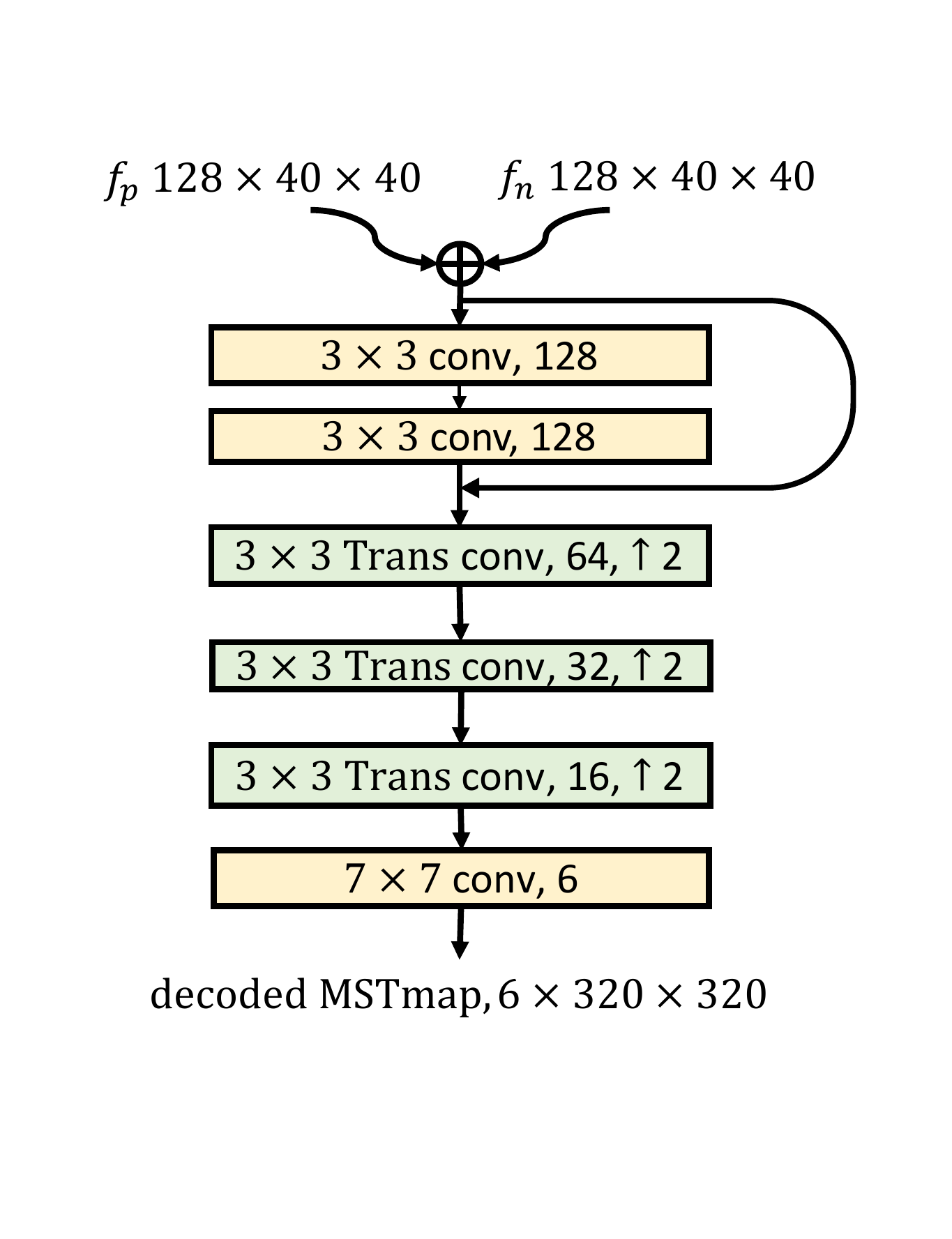}
}
\subfigure[]{
\includegraphics[width=0.36\linewidth]{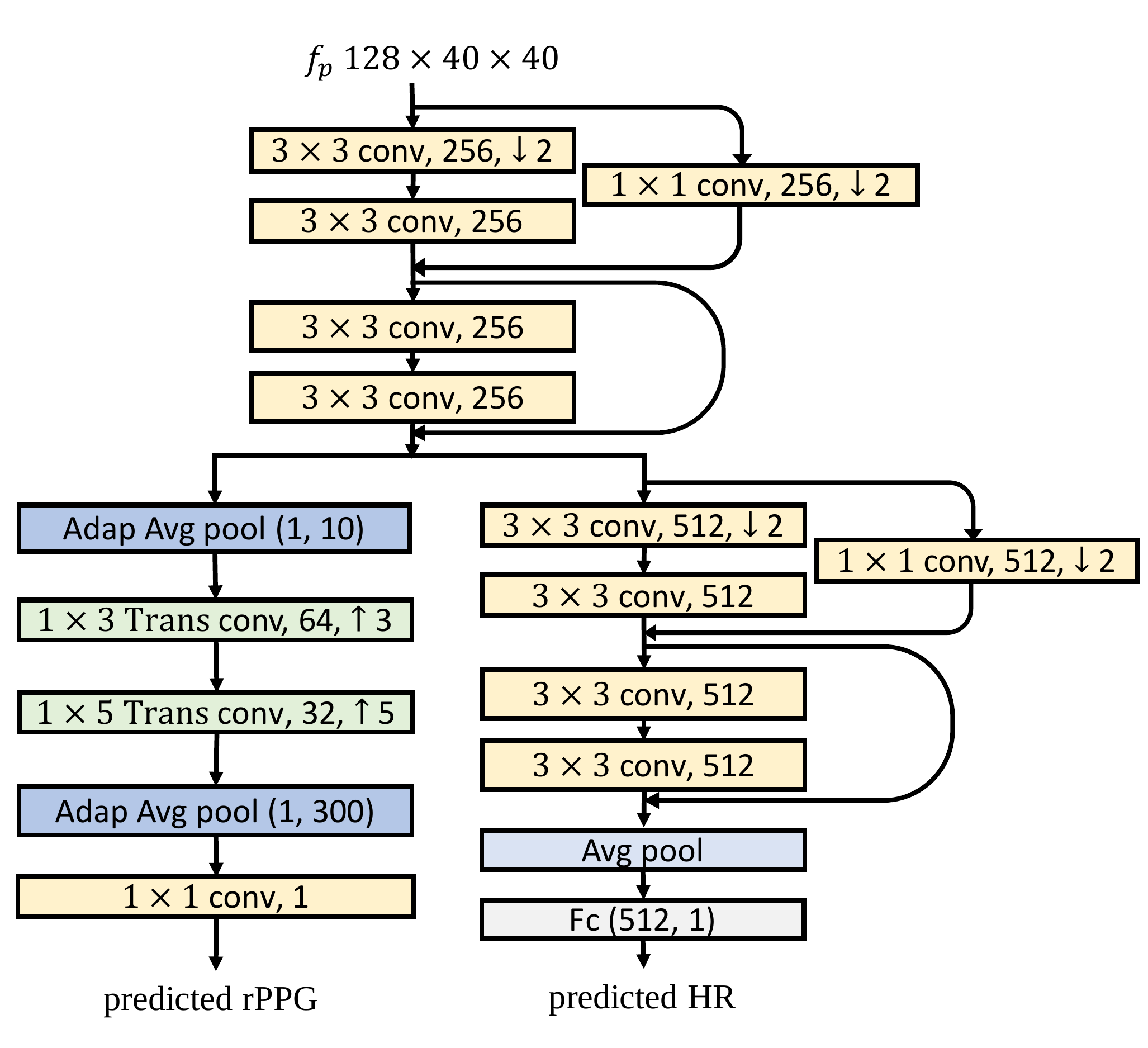}
\label{fig:estimator}
}
\caption{Detailed architectures of (a) the physiological and non-physiological encoder $E_p$ and $E_n$, (b) the decoder $D$, and (c) the physiological estimator. ``conv'' stands for convolution layers, and ``Trans conv'' represents transposed convolution layers. The number of output channels is denoted after the kernel size. ``$\downarrow$ and $\uparrow$'' mean down-sampling and up-sampling the feature map by setting the stride. ``Avg pool'' means average pooling layer, and ``Adap Avg pool'' represents the adaptive average pooling with its output size denoted behind. ``FC'' is fully-connected layer. We use batch normalization after every convolution layer in $E_p$, $E_n$ as well as the physiological estimator, and instance normalization after every convolution layer in $D$.
}
\label{fig:detail_network}
\end{figure}

\bibliographystyle{splncs04}
\bibliography{egbib}

\begin{thebibliography}{10}
\providecommand{\url}[1]{\texttt{#1}}
\providecommand{\urlprefix}{URL }
\providecommand{\doi}[1]{https://doi.org/#1}

\bibitem{carreira2017quo}
Carreira~J, Z.A.: Quo vadis, action recognition? a new model and the kinetics
  dataset. In: Proc. IEEE CVPR (2017)

\bibitem{chen2018deepphys}
Chen, W., Mcduff, D.: Deepphys: Video-based physiological measurement using
  convolutional attention networks. In: Proc. ECCV (2018)

\bibitem{de2013robust}
De~Haan, G., Jeanne, V.: Robust pulse rate from chrominance-based rppg. IEEE
  Trans. Biomed. Eng.  \textbf{60}(10),  2878--2886 (2013)

\bibitem{kingma2014adam}
Kingma, D.P., Ba, J.: Adam: A method for stochastic optimization. arXiv
  preprint arXiv:1412.6980  (2014)

\bibitem{Lam2015Robust}
Lam, A., Kuno, Y.: Robust heart rate measurement from video using select random
  patches. In: Proc. IEEE ICCV (2015)

\bibitem{Lee2019diverse}
Lee, H.Y., Tseng, H.Y., Huang, J.B., Singh, M., Yang, M.H.: Diverse
  image-to-image translation via disentangled representations. In: Proc. ECCV
  (2018)

\bibitem{lewandowska2011measuring}
Lewandowska, M., Ruminski, J., Kocejko, T., Nowak, J.: Measuring pulse rate
  with a webcam - a non-contact method for evaluating cardiac activity. In:
  Proc. ComSIS (2011)

\bibitem{li2018obf}
Li, X., Alikhani, I., Shi, J., Seppanen, T., Junttila, J., Majamaa-Voltti, K.,
  Tulppo, M., Zhao, G.: The {OBF} database: A large face video database for
  remote physiological signal measurement and atrial fibrillation detection.
  In: Proc. IEEE FG (2018)

\bibitem{li2014remote}
Li, X., Chen, J., Zhao, G., Pietikainen, M.: Remote heart rate measurement from
  face videos under realistic situations. In: Proc. IEEE CVPR (2014)

\bibitem{yu2018exploring}
Liu, Y., Wei, F., Shao, J., Sheng, L., Yan, J., Wang, X.: Exploring
  disentangled feature representation beyond face identification. In: Proc.
  IEEE CVPR (2018)

\bibitem{lu2019unsupervised}
Lu, B., Chen, J.C., Chellappa, R.: Unsupervised domain-specific deblurring via
  disentangled representations. In: Proc. IEEE CVPR (2019)

\bibitem{niu2018VIPL-HR}
Niu, X., Han, H., Shan, S., Chen, X.: {VIPL-HR}: A multi-modal database for
  pulse estimation from less-constrained face video. In: Proc. ACCV (2018)

\bibitem{niu2019rhythmnet}
Niu, X., Shan, S., Han, H., Chen, X.: Rhythmnet: End-to-end heart rate
  estimation from face via spatial-temporal representation. IEEE Trans. Image
  Processing  \textbf{29},  2409--2423 (2020)

\bibitem{niu2019robust}
Niu, X., Zhao, X., Han, H., Das, A., Dantcheva, A., Shan, S., Chen, X.: Robust
  remote heart rate estimation from face utilizing spatial-temporal attention.
  In: Proc. IEEE FG (2019)

\bibitem{poh2010non}
Poh, M.Z., McDuff, D.J., Picard, R.W.: Non-contact, automated cardiac pulse
  measurements using video imaging and blind source separation. Opt. Express
  \textbf{18}(10),  10762--10774 (2010)

\bibitem{poh2011advancements}
Poh, M.Z., McDuff, D.J., Picard, R.W.: Advancements in noncontact,
  multiparameter physiological measurements using a webcam. IEEE Trans. Biomed.
  Eng.  \textbf{58}(1),  7--11 (2011)

\bibitem{spetlik2018BMVC}
R., S., V., F., J., C., J., M.: Visual heart rate estimation with convolutional
  neural network. In: Proc. BMVC (2018)

\bibitem{luan2017disentangled}
Tran, L., Yin, X., Liu, X.: Disentangled representation learning {GAN} for
  pose-invariant face recognition. In: Proc. IEEE CVPR (2017)

\bibitem{Tulyakov2016Self}
Tulyakov, S., Alameda-Pineda, X., Ricci, E., Yin, L., Cohn, J.F., Sebe, N.:
  Self-adaptive matrix completion for heart rate estimation from face videos
  under realistic conditions. In: Proc. IEEE CVPR (2016)

\bibitem{verkruysse2008remote}
Verkruysse, W., Svaasand, L.O., Nelson, J.S.: Remote plethysmographic imaging
  using ambient light. Opt. Express  \textbf{16}(26),  21434--21445 (2008)

\bibitem{wang2019Discriminative}
Wang, W., den Brinker, A.C., de~Haan, G.: Discriminative signatures for
  remote-ppg. IEEE Trans. Biomed. Eng.  \textbf{67}(5),  1462--1473 (2020)

\bibitem{wang2017algorithmic}
Wang, W., den Brinker, A.C., Stuijk, S., de~Haan, G.: Algorithmic principles of
  remote ppg. IEEE Trans. Biomed. Eng.  \textbf{64}(7),  1479--1491 (2017)

\bibitem{wang2017amplitude}
Wang, W., den Brinker, A.C., Stuijk, S., de~Haan, G.: Amplitude-selective
  filtering for remote-ppg. Biomed. Opt. Express  \textbf{8}(3),  1965--1980
  (2017)

\bibitem{wang2015exploiting}
Wang, W., Stuijk, S., De~Haan, G.: Exploiting spatial redundancy of image
  sensor for motion robust rppg. IEEE Trans. Biomed. Eng.  \textbf{62}(2),
  415--425 (2015)

\bibitem{yu2019ICCV}
Yu, Z., Peng, W., Li, X., Hong, X., Zhao, G.: Remote heart rate measurement
  from highly compressed facial videos: An end-to-end deep learning solution
  with video enhancement. In: Proc. IEEE ICCV (2019)

\bibitem{zhang2019gait}
Zhang, Z., Tran, L., Yin, X., Atoum, Y., Liu, X.: Gait recognition via
  disentangled representation learning. In: Proc. IEEE CVPR (2019)

\end{thebibliography}

\end{document}